\documentclass[conference]{IEEEtran}
\IEEEoverridecommandlockouts

\usepackage{cite}
\usepackage{amsmath,amssymb,amsfonts}
\usepackage{algorithmic}
\usepackage{graphicx}
\usepackage{svg}
\usepackage{textcomp}
\usepackage{xcolor}
\usepackage{multirow}
\usepackage{caption}
\usepackage{booktabs}
\usepackage{array}
\usepackage{stfloats}
\usepackage[hidelinks]{hyperref} 
\def\BibTeX{{\rm B\kern-.05em{\sc i\kern-.025em b}\kern-.08em
    T\kern-.1667em\lower.7ex\hbox{E}\kern-.125emX}}
    
\begin{document}

\title{Benchmarking Energy and Latency in TinyML: \\ A Novel Method for Resource-Constrained AI}

\author{
\IEEEauthorblockN{
Pietro Bartoli\textsuperscript{$^1$}\IEEEauthorrefmark{1}\thanks{\textsuperscript{$1$}These authors contributed equally to this work.}, 
Christian  Veronesi\textsuperscript{$^1$}\IEEEauthorrefmark{1},
Andrea Giudici\IEEEauthorrefmark{1},
David Siorpaes\IEEEauthorrefmark{2},
Diana
Trojaniello\IEEEauthorrefmark{3}
Franco Zappa\IEEEauthorrefmark{1},
}
\IEEEauthorblockA{\textit{\IEEEauthorrefmark{1}Dept. of Electronics, Information and Bioengineering, Politecnico di Milano, Milan, Italy}}
\IEEEauthorblockA{\textit{\IEEEauthorrefmark{2}STMicroelectornics, Agrate Brianza (MB), Italy}}
\IEEEauthorblockA{\textit{\IEEEauthorrefmark{3}Smart Eyewear Laboratory, EssilorLuxottica, Milan, Italy}}
}

\maketitle

\begin{abstract}
The rise of IoT has increased the need for on-edge machine learning, with tinyML emerging as a promising solution for resource-constrained devices such as Microcontroller Units (MCUs). 
However, benchmarking their energy efficiency, latency, and computational capabilities remains challenging due to diverse architectures and application scenarios. 
Current solutions, such as MLCommons’ "TinyML Perf: Inference" method, have limitations, including the need for separate setups for latency, accuracy, and energy measurements, as well as reliance on energy monitors that power the device, reducing flexibility. 
Moreover, the absence of a clear distinction between inference and ancillary operations can compromise the accuracy of performance estimations.
This work introduces an alternative benchmarking methodology that integrates energy and latency measurements while distinguishing three execution phases—pre-inference, inference, and post-inference—to enable precise profiling.
A dual-trigger approach is used to separate these phases, ensuring accurate and repeatable measurements. 
Additionally, the setup ensures that the device operates without being powered by an external measurement unit, while automated testing can be leveraged to enhance statistical significance.
To evaluate our setup, we tested the STM32N6 MCU, which includes a Neural Processing Unit (NPU) for executing convolutional neural networks. 
Two configurations were considered: one running at maximum performance with the highest clock frequencies and core voltage, and another with reduced settings. 
The variation of the Energy Delay Product (EDP) was analyzed separately for each phase, providing insights into the impact of hardware configurations on energy efficiency. 
Each MLPerf model was tested 1000 times to ensure statistically relevant results.
Our findings demonstrate that reducing the core voltage and clock frequency improves the efficiency of pre- and post-processing without significantly affecting network execution performance.
This approach can also be used for cross-platform comparisons to determine the most efficient inference platform and to quantify how pre- and post-processing overhead varies across different hardware implementations.
\end{abstract}


\section{Introduction}
The rapid expansion of IoT has created a globally interconnected network of billions of devices, primarily relying on cloud computing for intelligent and autonomous operations~\cite{Atzori_2010, Ali_2015}.
However, this dependence introduces latency, security risks, and high energy consumption due to the need for offloading computations~\cite{Sipola_2022, Singh_2023}.

Tiny Machine Learning (TinyML) addresses these challenges by enabling local machine learning (ML) inference on low-power MCUs, reducing latency and energy overhead~\cite{Tsoukas_2024, Doyu_2021}. 
However, MCUs’ limited memory and computational power make efficient neural network execution non-trivial~\cite{Adlakha_2024}.

To overcome these constraints, Neural Processing Units (NPUs) have emerged as a promising solution, offloading ML tasks to dedicated hardware optimized for deep learning~\cite{Cardarilli_2018}. 

As TinyML systems grow more complex with diverse hardware and NPUs, a universal benchmarking framework is essential for fair comparisons and optimizations~\cite{Giordano_2022}. 
In particular, NPUs can be coupled with the MCU at different layers adding variability, and making it crucial to isolate their contribution, especially when external, to assess energy efficiency and inference latency accurately.
However, isolating only the inference execution is not yet a common practice, making it difficult to fully understand its impact. This approach is essential for evaluating the potential benefits of an NPU compared to a traditional CPU and determining whether its use is advantageous in a given scenario.

\section{Benchmarking MCUs}

As previously said, evaluating MCU performance is essential due to the diverse applications of ML models, ranging from Industry 4.0 systems~\cite{Casiroli_2023,Antonini_2022,Tsoukas_2023} to wearable health-monitoring devices~\cite{Tsoukas_2022,Ahmed_2022}.
Indeed, each application has unique requirements in execution time, accuracy of results, ML model complexity, and energy consumption.
However, the wide variety of existing MCUs, each with unique features, capabilities, and programming environments~\cite{Banbury_2021,vanKempen_2024}, makes identifying the most suitable option a complex task.

A systematic, repeatable, and robust benchmarking methodology is required to address these challenges and rigorously assess the benefits of NPU integration.
In particular, benchmarking must primarily focus on three key metrics: accuracy, latency, and energy consumption, since these metrics are essential for evaluating both the performance of the model and the efficiency of the hardware it runs on.

To meet the specific demands of TinyML, different benchmarking frameworks have been introduced; the most popular one is MLPerf Tiny developed by MLCommons~\cite{Banbury_2021_MLPerf}, described in the following section.
Despite the availability of this framework, ongoing research continues to focus on improving benchmarking processes and refining the datasets used for the evaluation process~\cite{Banbury_2024}.

\subsection{MLCommons Benchmarking Suite}
In this section, we briefly describe the MLPerf Tiny benchmarking procedure to provide a context for identifying its limitations.
MLPerf Tiny is recognized as a standard for performance evaluation in the TinyML ecosystem~\cite{Reddi2020}. 
The benchmarking setup consists of three primary components:

\begin{itemize}
    \item \textbf{Host PC}: It manages the benchmarking procedure through the runner application.
    \item \textbf{Energy Monitor}: Measurement board, e.g., LPM01A by ST-Microelectronics.
    \item \textbf{Device Under Test} (DUT): System including the MCU under test.
\end{itemize} 

The benchmark evaluates TinyML systems across four key tasks, summarized in Table \ref{tab:tasks}. 

\begin{table}[h]
\caption{Overview of tasks, datasets, and models selected by MLCommons for MLPerf Tiny benchmark suite.}
\resizebox{\columnwidth}{!}{ 
\begin{tabular}{l|l|c}
\multicolumn{1}{c|}{\textbf{Task}} & \multicolumn{1}{c|}{\textbf{Dataset}} & \textbf{Model}   \\ \hline
\textit{Keyword Spotting}          & Google Speech Commands                & DS-CNN           \\ \hline
\textit{Visual Wake Word}          & Visual Wake Word Dataset              & MobileNet        \\ \hline
\textit{Image Classification}      & CIFAR10                               & ResNet-8         \\ \hline
\textit{Anomaly Detection}         & Toy Admos                             & Deep Autoencoder
\end{tabular}
}
\label{tab:tasks}
\end{table}

The MLPerf Tiny benchmark suite evaluates TinyML systems using the three aforementioned metrics: latency, accuracy, and energy consumption. 

\textit{Latency} is measured by testing inference speed through repeated runs on standard inputs, with results reported in terms of inferences per second. 

\textit{Accuracy} is assessed by comparing system outputs against predefined thresholds on validation datasets, which must be met by participants to qualify for benchmarking. 

\textit{Energy consumption} is measured during the inference and is given as $\mu J$ per inference. 

These measurements are performed through a dedicated benchmarking setup designed to coordinate the interactions between the test environment and the DUT. 

Latency measurement begins by loading a predefined input stimulus onto the DUT. The inference process is executed for a minimum duration of 10 seconds and at least 10 iterations.
During each run, the number of inferences performed per second is recorded. 
This procedure is repeated five times, and the median value of inferences per second is reported as the latency score.

Energy measurements follow a similar procedure, with additional instrumentation integrated into the setup. 
The energy monitor is connected to the DUT to measure the total energy consumed during the inference phase. 
As for latency, the measurement is repeated five times, and the median energy consumption per inference is reported.

The framework is configured in two distinct setups to address specific measurement objectives.
For latency and accuracy measurements, the DUT is connected directly to the host PC via a serial port, enabling communication and control.
For energy measurements, an energy meter powers the DUT while isolating the energy consumption of the DUT core running the inference, excluding contributions from peripherals and framework overhead.

Both measurement setups follow the same process for loading input data, initiating inference, and collecting results, guaranteeing consistency in methodology.
Its modular design enables the evaluation of essential tasks requiring different network architectures and data types.
This framework facilitates the evaluation of the energy consumed by the DUT.

However, it presents several major limitations:
\begin{itemize} 
    \item \textbf{Impact of Non-Inference Phases}: 
    Energy consumption and latency measurements are always attributed to the inference process; however, they also encompass contributions from pre- and post-inference phases, leading to inaccuracies in the evaluation. 
    This limitation arises from the use of a single trigger signal to initiate the measurement process, which fails to isolate the inference phase from surrounding operations. 
    Consequently, the inclusion of non-inference activities within the measured results skews the evaluation, resulting in an underestimation of the actual energy consumed solely by inference~\cite{PAUarticle}.
    \item \textbf{Absence of Energy-Latency-Accuracy Consistency}: 
    The current protocol exploits two distinct measurement setups for energy and accuracy, and latency evaluation. 
    Moreover, energy consumption and latency are measured over 4 cycles~\cite{PAUarticle}, while the accuracy is computed on an entire set of validation inputs. 
    This methodological discrepancy precludes the simultaneous execution of all the measurements.
    The final results are reported without accounting for measurement uncertainty, reducing their reliability, and the limited number of results prevents the derivation of meaningful statistical insights.
    \item \textbf{Need for Energy-Latency Correlation}:
    Sometimes, comparing performance between different devices and identifying the best one could be difficult. Considering battery-powered IoT devices, the trade-off between energy and latency is a crucial metric. As a result, providing unique metrics that measure this trade-off provides a direct comparison between different MCUs. 
    \item \textbf{Specific Power Requirements}: 
    The DUT must be powered directly by the energy monitoring equipment, which may limit flexibility. 
    As a consequence, when the required core voltage is outside the possible range provided by the equipment, additional electronics are mandatory to adapt the voltage, introducing additional contributions to energy. 
    For example, the STMicroelectronics LPM01A energy meter is able to provide a voltage from $1.8\text{V}$ to $3.3\text{V}$. 
    However, cores of MCUs built using nanometric technologies often require core voltages below $1\text{V}$.
\end{itemize}

\section{New Benchmarking Methodology}

\begin{figure}[b]
     \centering
     \captionsetup{font=small}
     \includegraphics[width=0.45\textwidth]{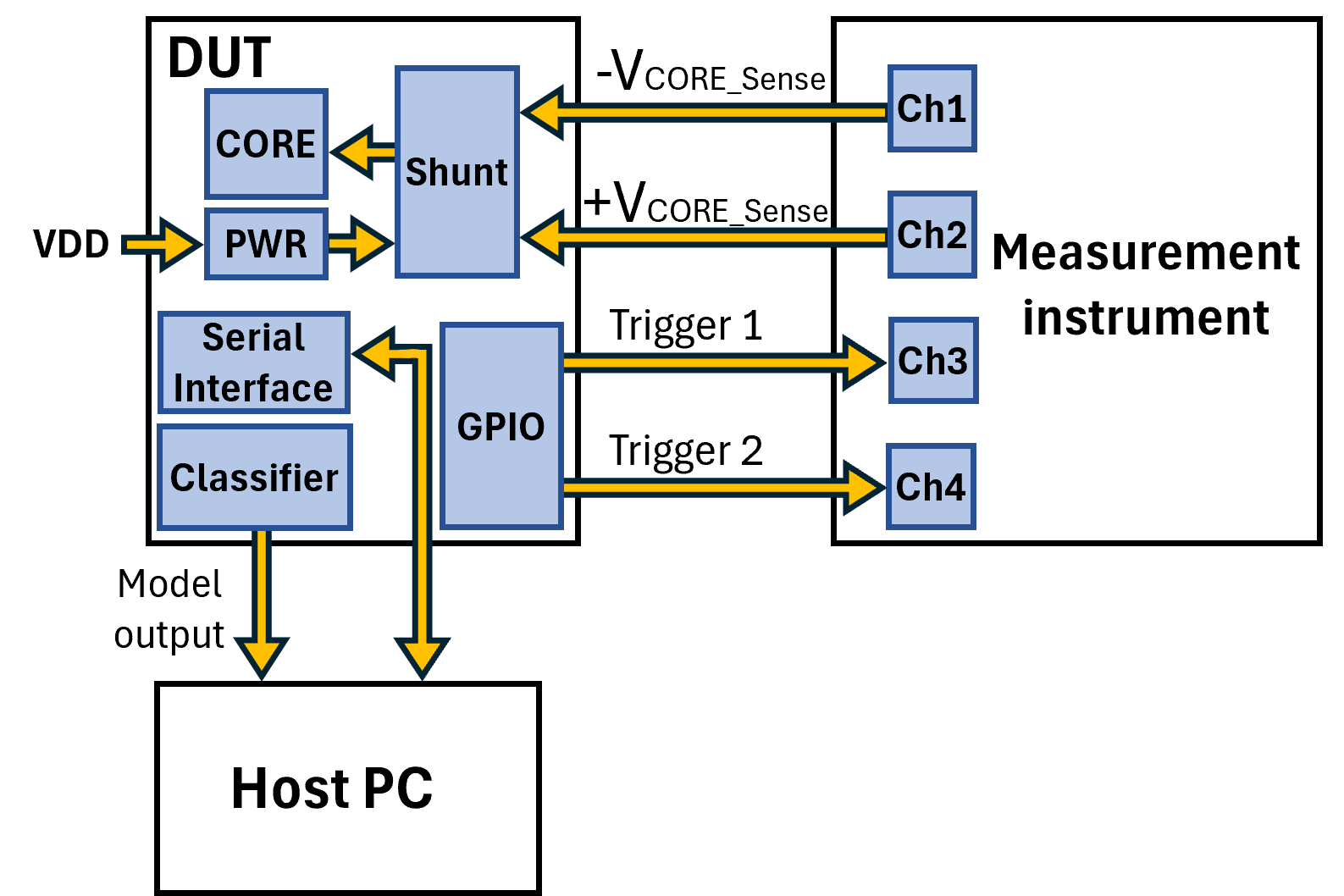}
     \caption{Block diagram of the measurement setup. The measurement instrument measures the voltage drop across the shunt resistor from \textit{Ch1} and \textit{Ch2}, and monitors two trigger signals from \textit{Ch3} and \textit{Ch4} to distinguish four different phases (Pre-, Post-, inference, and idle). The PC communicates with the DUT through the serial interface. Meanwhile, the MCU automatically transmits classifier outputs to the Host PC, enabling accuracy measurements during a subsequent phase and setting or resetting one of the reference signals during each phase} \label{fig:Scheme_setup}
\end{figure}

To overcome these problems, we developed a new benchmarking methodology with the ultimate goal of providing more accurate and precise results while overcoming all the limitations of the MLPerf Tiny method.
Our proposed benchmarking methodology is based on measuring the energy consumption directly on the DUT through a shunt resistor placed in series to the MCU's core domain.
The setup reported in Figure~\ref{fig:Scheme_setup} is based on three main blocks:

\begin{itemize}
    \item {\bf DUT}: The target board that hosts the high-precision shunt resistor and the MCU to benchmark, which executes specific Firmware (FW) tailored for benchmarking.
    \item {\bf Bench Measurement Instrument}: It acquires the voltage drop across the shunt resistor and two trigger signals provided by the DUT.
    \item {\bf Host PC}: It acquires the DUT output prediction to compute the model accuracy.
\end{itemize}

In battery-powered IoT devices, evaluating energy consumption is critical for estimating system autonomy. 
To address this, we used a double-trigger signal to measure both the energy consumption and the time required for each operational phase.
We identify four different phases (Pre-Inference, Inference, Post-Inference, and Idle) defining four different signal windows exploiting two GPIO-based signals.
Before each single phase, one signal is changed to indicate the finish of the actual phase and the beginning of the next one (see Table~\ref{tab:ref_signals} for the signal coding configurations) avoiding any possible mismatch artifacts.
The rise and fall times of the trigger signals are negligible compared to the duration of the measured phases, ensuring they do not affect the accuracy of the measurements.
The use of GPIO-based signaling for precise timing in TinyML systems aligns with methodologies established for reliable energy and latency measurements in such contexts~\cite{Reddi_24}. 

\begin{table}[h]
    \centering
    \renewcommand{\arraystretch}{1.2}
    \caption{Signal coding steps}
    \resizebox{0.3\textwidth}{!}{%
    \begin{tabular}{c|cc}
        \hline
        \bf{Phase}             & \textit{Trigger 1} & \textit{Trigger 2} \\
        \hline
        \textit{Pre-Inference} & 1                  & 0 \\
        \textit{Inference}     & 1                  & 1 \\
        \textit{Pre-Inference} & 0                  & 1 \\
        \textit{Idle}          & 0                  & 0 \\
        \hline
    \end{tabular}%
    }
    \label{tab:ref_signals}
\end{table}
 
This approach ensures very precise phase isolation while maintaining synchronization integrity, and it helps pinpoint specific phases where optimization efforts can be focused, offering detailed insights into the energy and performance impact of each stage.

Energy consumption is measured by recording the voltage drop across a shunt resistor, $R_{shunt}$, connected in series to the DUT core, which allows the calculation of the current absorbed, $I_{core}$.
In our setup, $R_{shunt}$ has a value of $50 \text{m}\Omega$ with a tolerance of $\pm 1\%$, providing sufficient Signal-to-Noise Ratio (SNR) for accurate readings.
Power consumption is then determined by multiplying the core current by the core voltage, and energy consumption is obtained by integrating the power over time. 
The dual-trigger approach ensures these calculations can be independently performed for all three phases previously identified, enabling a more granular understanding of energy usage.

To streamline this process, three equations are defined, Eq.~\ref{eq:current},\ref{eq:power},\ref{eq:energy}:

\begin{align}
    I_{\text{core}}\left(t\right) = \frac{\Delta V_{shunt}}{R_{shunt}} 
    \label{eq:current}
\end{align}
\begin{align}
    P_{\text{core}}\left(t\right) = \displaystyle I_{\text{core}}\left(t\right) \cdot V_{\text{core}} 
    \label{eq:power}
\end{align}
\begin{align}
    \overline{E}_{core} = \int_{t_0}^{t_1} \displaystyle P_{\text{core}}\left(t\right) \cdot  d t
    \label{eq:energy}
\end{align}

In equation~\ref{eq:energy}, the $\overline{E}_{core}$ is obtained by integrating the $P_{core}$ between $t_0$ and $t_1$ which are the time instants identified by the edges of the considered window.  
This methodology ensures an accurate evaluation of energy consumption for each phase, separating all contributions.
This benchmarking methodology supports an adjustable number of measurements, allowing enhanced statistical significance and ensuring more reliable and robust performance evaluation, compared to the methodology defined by MLCommons~\cite{Reddi2020}.
The measurement process can be summarized in the following nine steps:

\begin{enumerate}
     \item \textbf{Upload and Run the Firmware}: The process begins by uploading the FW onto the DUT and initiating its execution to prepare the device for the measurement procedure.
    
     \item \textbf{Measurement instrument Configuration}: The measurement instrument is configured to set up all necessary measurement options, so to ensure that the equipment is ready to capture the required data accurately during the measurement process.

     \item \textbf{Measurement acquisition}: When all systems are ready the measurement process starts, and the measurement instrument begins to acquire both analog and digital signals coming from the DUT.

     \item \textbf{Data interpolation}: Once the measurement process is completed, the data can be evaluated. The two trigger signals are interpolated to generate three independent consecutive windows following Table~\ref{tab:ref_signals}. From these windows, it is possible to measure the duration of each phase directly. The energy consumption of each phase is obtained through Eq~\ref{eq:energy}, limiting the integrating interval to the considered window.

     \item \textbf{Final results}: The results of each single acquisition are then averaged, obtaining the mean energy consumption and latency of each single phase. Moreover, to increase the result's robustness, all standard deviations are computed, providing information regarding possible performance deviation of the DUT.  
    
\end{enumerate}

\subsection{Energy Delay Product}

Our proposed benchmarking methodology prescribes the use of the Energy-Delay Product ($EDP$) as a key metric for evaluating efficiency. 
We have chosen $EDP$ because it effectively captures the trade-off between energy consumption and execution time, making it particularly suitable for assessing ML inference on MCUs.

The $EDP$ is a widely recognized metric in the literature and is commonly used in latency-sensitive applications to quantify efficiency~\cite{Garcia_2019}.
By combining energy and performance into a single value, $EDP$ could simplify comparisons across different configurations and platforms, offering a concise and meaningful measure of efficiency.
The $EDP$ is defined as~\ref{eq:EDP}:

\begin{equation} 
EDP = E_{i} \cdot T_{i} 
\label{eq:EDP} 
\end{equation}

where $E_{i}$ and $T_{i}$ represent the energy consumption and latency of the $i$-th configuration, respectively.

Moreover, we chose to express $EDP$ as the \textit{Relative Energy-Delay Product} ($rEDP$) to present the results as percentage variations, compared to a reference configuration, in order to quantify the efficiency variation.

\subsection{Key Advantages}
The proposed methodology offers several key advantages over current state-of-the-art benchmarking processes.

Firstly, by placing a shunt resistor near the core, energy consumption can be measured directly at the core level, ensuring that the recorded energy primarily reflects the core’s contribution while minimizing interference. 

Additionally, eliminating the need for a dedicated energy meter allows the DUT to use its own power supply, enhancing flexibility. 
Otherwise, if the required core voltage is not provided by the measurement instrument, additional electronics would be needed for proper operation and accurate measurements.

Moreover, the use of two separate trigger signals generated by the DUT enables precise segmentation of execution phases.
These signals, processed through logic equations, create a unique binary encoding that clearly distinguishes pre-inference, inference, and post-inference stages. 
As a consequence, with $n$ trigger signals, this method can define up to $2^n$ execution states, enabling flexibility in adapting to different benchmarking needs, for example, the execution of two consecutive models.
In contrast, the state-of-the-art relies on a single trigger signal to synchronize acquisition, which fails to distinguish between different execution phases. 
This limitation prevents fair comparisons when evaluating NPUs and makes optimization more challenging, as it obscures the contribution of individual stages to overall performance.

Furthermore, this methodology allows for rapid repetition of measurements, enabling the collection of a large dataset where both the mean and standard deviation can be reliably analyzed.
This allows the selection of the number of inferences to be executed, with the possibility of extending the benchmark to the whole dataset, enabling the measurement of accuracy, latency, and energy consumption from the same test.
By consolidating all measurements into a single setup, this approach ensures that results are both representative and statistically robust while reducing the complexity and variability associated with multiple experimental configurations.

Lastly, the usage of the $EDP$ allows to express the energy efficiency of the DUT using a single value, describing more intuitively the trade-off between these two quantities.
On the other hand, the $rEDP$ is useful to compare either different setups of the same device, or different devices, allowing faster and easier comparison.

\section{Benchmark Test}
To validate the effectiveness of our benchmarking setup, we conducted measurements on the STM32N6, a cutting-edge MCU developed by STMicroelectronics. 
As the first STM32 product in a new series of MCUs based on the Cortex-M55 architecture~\cite{arm_M55}, the STM32N6 integrates a powerful convolutional NPU capable of achieving $0.6 \text{TOPS}$ at $1 \text{GHz}$. 
This flash-less device features up to 4.2 MBytes of internal RAM, providing flexibility in storing model weights and activations, which can be allocated across internal RAM, external Flash, and/or external RAM. 
Its highly configurable architecture makes it an excellent platform for demonstrating how our measurement setup combined with $rEDP$ can guide the selection of the optimal configuration, enabling a systematic optimization of the trade-off between performance and energy consumption in embedded AI applications.

\begin{figure}[h]
    \centering
    \includegraphics[width=0.38\textwidth]{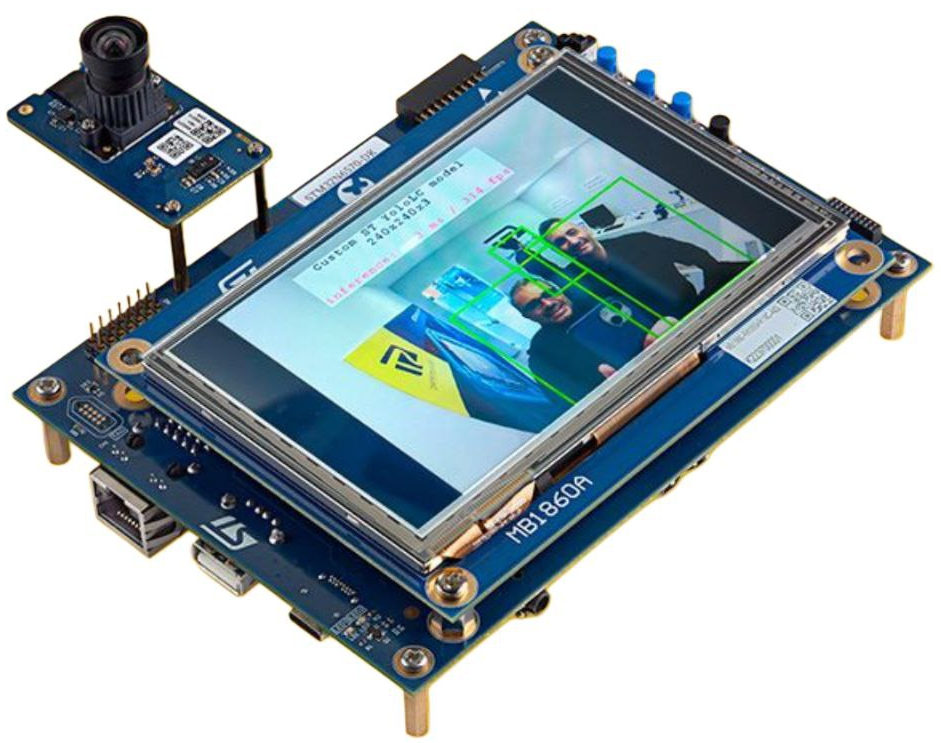}
    \captionsetup{font=small}
    \caption{Development kit STM32N6570-DK used for the measurements, featuring the N6 MCU and multiple peripherals}
    \label{fig:dk}
\end{figure}

We performed our measurements using the STM32N6570-DK development kit, which includes the N6 MCU along with various peripherals such as a microphone, cameras, and a display (Figure~\ref{fig:dk}).

Performance evaluation was conducted on the four networks defined by MLPerf under two configurations: High Performance (H-Perf) and Low Performance (L-Perf). 
The H-Perf mode, characterized by maximum core voltage and clock frequency, operates with increased settings for memory and the NPU, while L-Perf reduces these parameters to lower power consumption.
Table~\ref{tab:config} details the parameters for each.

For the comparative analysis, we used the $rEDP$, taking H-Perf as the reference, allowing to assess the impact of CPU speed, memory access, and NPU settings on energy efficiency.

\begin{table}[h]
    \centering
    \renewcommand{\arraystretch}{1.2}
    \caption{Voltage and frequency for H-Perf and L-Perf configurations.}
    \resizebox{0.48\textwidth}{!}{%
    \begin{tabular}{r|cccc}
        \hline
        \textbf{Configuration} & $V_{CORE}$ ($mV$) & $F_{CPU}$ ($MHz$) & $F_{NPU}$ ($MHz$) & $F_{RAM}$ ($MHz$) \\
        \hline
        \textit{H-Perf}       & 900  & 800  & 1000 & 900 \\
        \textit{L-Perf} & 800  & 800  & 800  & 720 \\
        \hline
    \end{tabular}%
    }
    \label{tab:config}
\end{table}

Specifically, the energy measurements were performed using the MSO64B oscilloscope by \textit{Tektronix} while the N6 was executing a specific firmware composed of four main steps:

\begin{enumerate}
    \item {\bf Initial setup (idle)}: This step is executed only once when the firmware is uploaded. During this phase, the N6 initializes all required peripherals and moves the model weights from the external flash to the internal RAMs. After this phase, the benchmarking process starts. 
    This section is not an object of measurement, since during this phase the DUT only performs the initial configuration before starting the benchmarking process. 
    \item {\bf Pre-Inference}: During this step, the N6 sets high the state of the Trigger 1 signal. Then it loads an input tensor from the external flash, previously stored, into the internal RAMs.
    \item {\bf Inference}: The N6 sets to high the state of the Trigger 2 signal before executing the model. During this phase, all unnecessary peripherals are clock-gated to reduce energy consumption.   
    \item {\bf Post-Inference}: The N6 resets the state of the Trigger 1 signal. Then, it executes the specific post-process model and sends the model prediction through serial interface to the PC, allowing the evaluation of the model's accuracy. 
    Subsequently, the procedure restarts from the Pre-Inference phase.
\end{enumerate}

\begin{figure}[]
    \centering
    \includegraphics[width=0.35\textwidth]{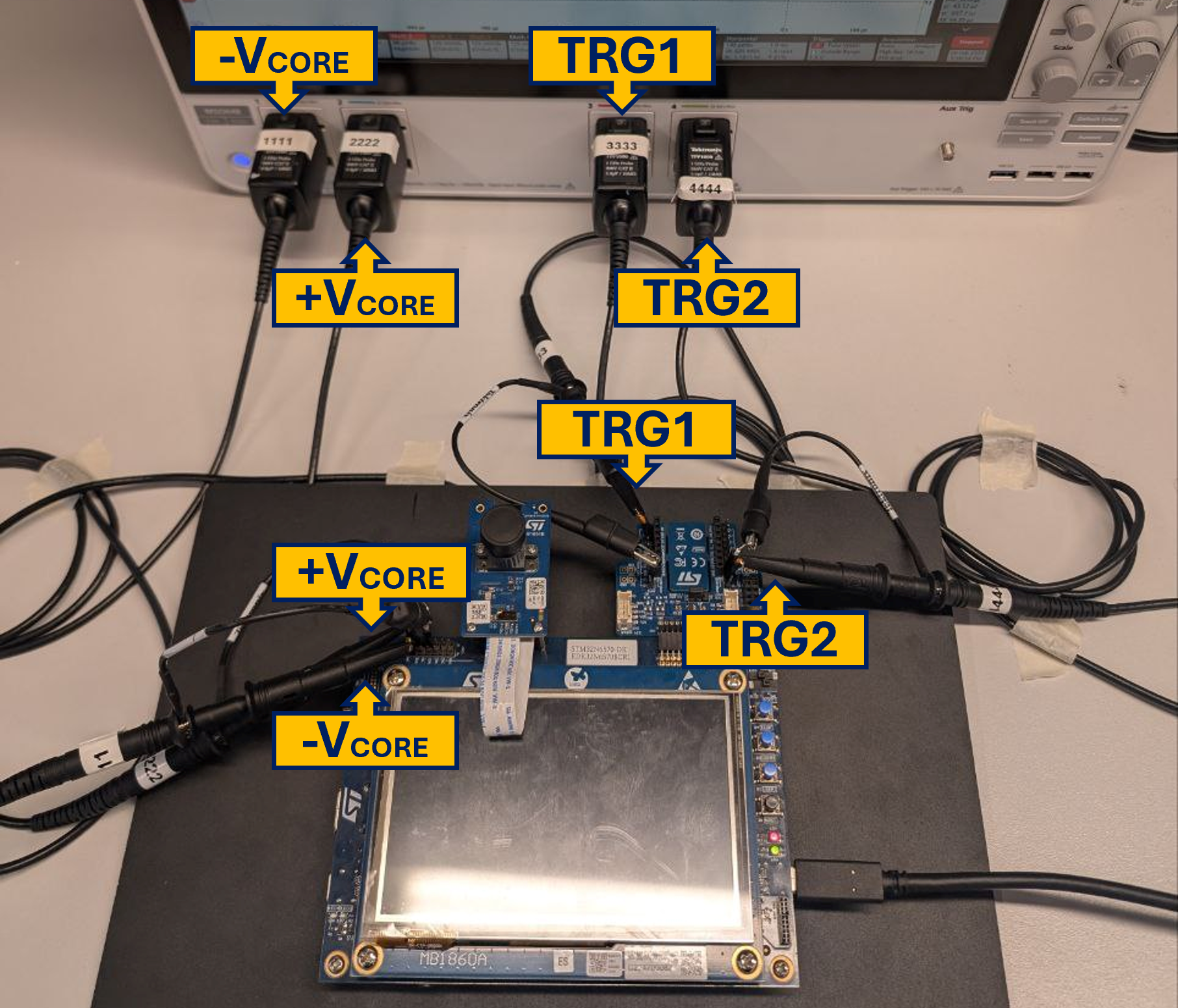}
    \captionsetup{font=small}
    \caption{Measurement setup showing the STM32N6570-DK development kit connected to the MSO64B oscilloscope. On the left, two probes measure the voltage drop across \( R_{shunt} \) of the STM32N6 core, while on the right, two probes detect the two trigger signals.}
    \label{fig:oscilloscope}
\end{figure}

Figure~\ref{fig:oscilloscope} illustrates the measurement setup, where the development kit is connected to the oscilloscope via four probes: two for measuring the voltage drop across $R_{shunt}$ of the STM32N6 core and two for detecting the trigger signals. 
We manage the interpolation of the two trigger signals by the oscilloscope defining three windows for each phase.

\begin{figure}[]
    \centering
    \includegraphics[width=0.48\textwidth]{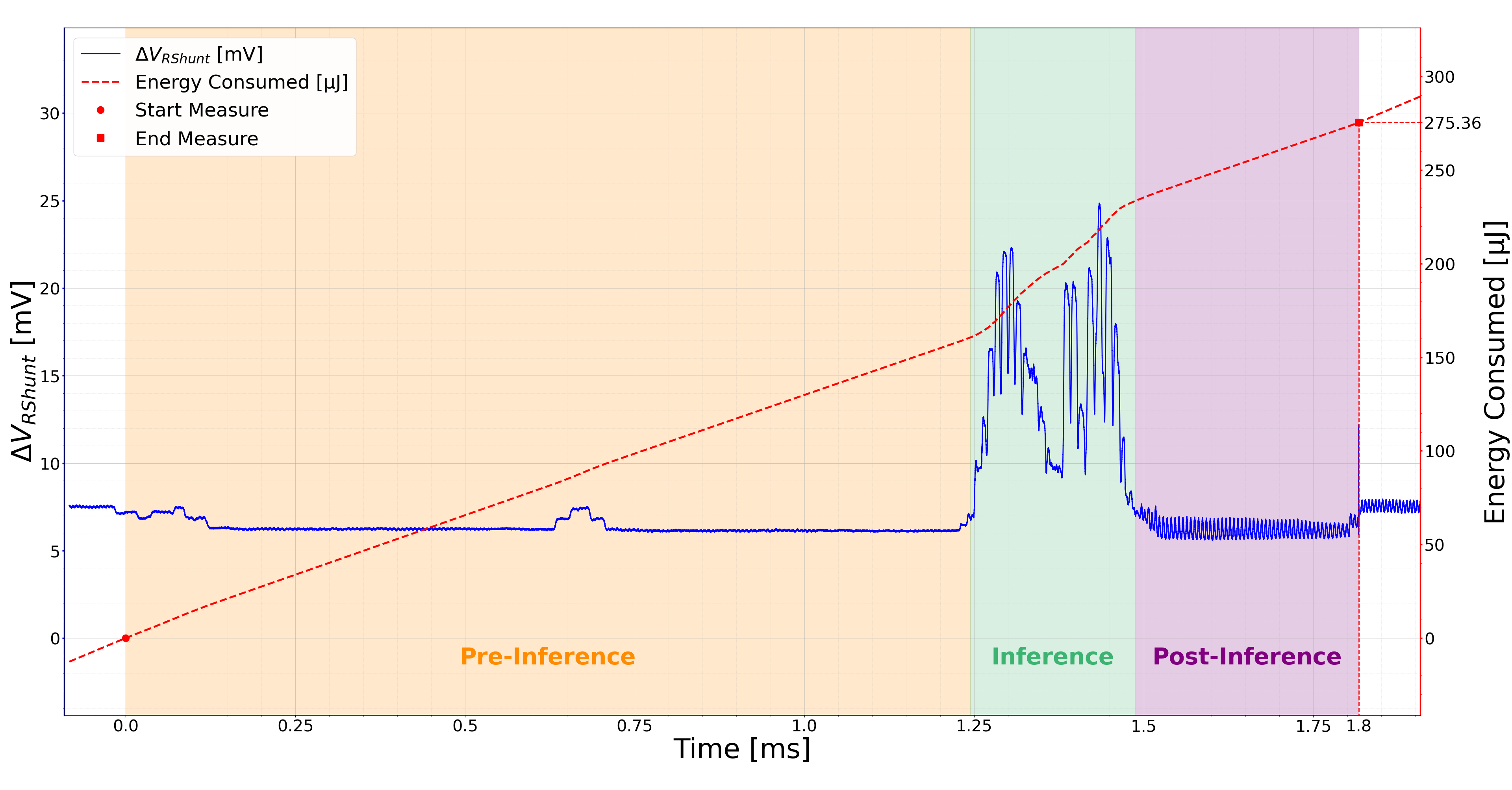}
    \captionsetup{font=small}
    \caption{Oscilloscope trace of the energy consumption during an inference process, segmented into \textit{Pre-Inference} (orange), \textit{Inference} (green), and \textit{Post-Inference} (purple) phases. 
    The blue trace shows the voltage drop across $R_\text{shunt}$, while the red dashed line indicates cumulative energy consumption. 
    Circular and square markers denote the start and end of the energy measurement interval, respectively.}
    \label{fig:screen}
\end{figure}

Figure~\ref{fig:screen} shows the system's behavior across the three phases, where the distinct energy consumption patterns in each phase can be identified thanks to the windowing signals.

\section{Results and Discussion}
In this section, we discuss the performance of the STM32N6 obtained through our benchmarking setup. 

Firstly, we conducted measurements under the H-Perf configuration. 
Tables~\ref{tab:energy} and \ref{tab:time} show the energy consumption and latency for the three execution phases averaged over 1000 acquisitions. 
The results indicate that the pre-inference phase contributes the most to energy consumption and latency. 
Notably, during pre-inference, data is stored in the internal RAM, in preparation for neural network execution, making this phase a necessary step in the processing pipeline.
L-Perf was introduced to optimize key system parameters, minimizing energy usage during the pre-inference phase, which is handled by the MCU’s CPU.
Specifically, we lowered the core voltage from $900 mV$ to $800 mV$ and reduced the clock frequency of both the NPU and its dedicated RAMs, as these components always provide a lower contribution in both energy and latency. 
However, the CPU clock frequency remained unchanged to preserve the pre- and post-inference latency, ensuring a balance between energy savings and computational efficiency.
This results in a slight increase in the model latency, as reported in Table~\ref{tab:time}.
However, this increment in latency is almost negligible when the total latency is considered.
On the contrary, the energy consumption decreases for all three phases when moving to L-Perf configuration, leading to a significant overall energy consumption reduction as shown in Table~\ref{tab:energy}.
The main reason comes from the core voltage reduction that can be achieved thanks to a lower frequency for the NPU and its RAMs.

\begin{figure*}[]
    \centering
    \includegraphics[width=0.95\textwidth,height=0.25\textheight]{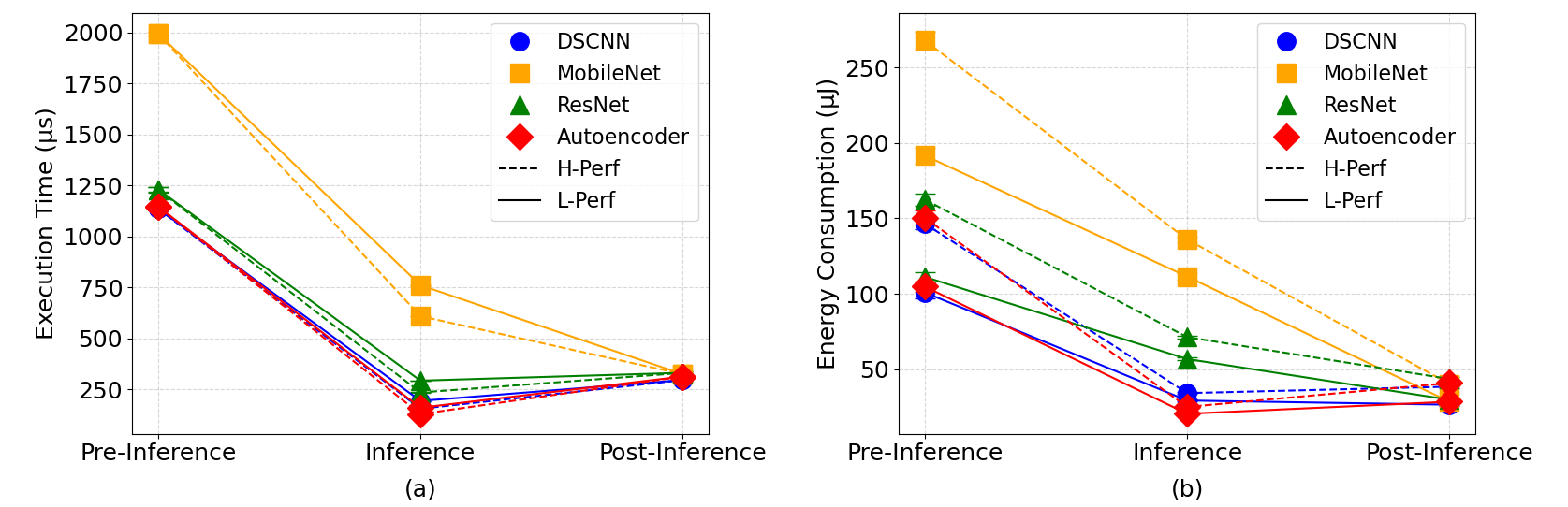}
    \captionsetup{belowskip=-10pt} 
    \caption{Comparison of execution time (a) and energy consumption (b) across the three phases under H-Perf (dashed line) and L-Perf (continuous lines) configurations. The execution time (latency) is the same for both pre- and post- inference phases since the CPU is speed is the same, while there is a reduction in energy consumption primarily due to voltage core reduction}
    \vspace{10pt}
    \label{fig:line}
\end{figure*}

Figure~\ref{fig:line} shows these trends, which highlight the variations in execution time and energy consumption across the three phases for both configurations.

To quantify the energy efficiency gained with the L-Perf configuration, we first computed the EDP using the mean values over 1000 executions. 
We then calculated the rEDP when transitioning from H-Perf to L-Perf. 
The results of this analysis are reported in Table~\ref{tab:redp}.

\begin{figure*}[]
    \centering    \includegraphics[width=\textwidth]{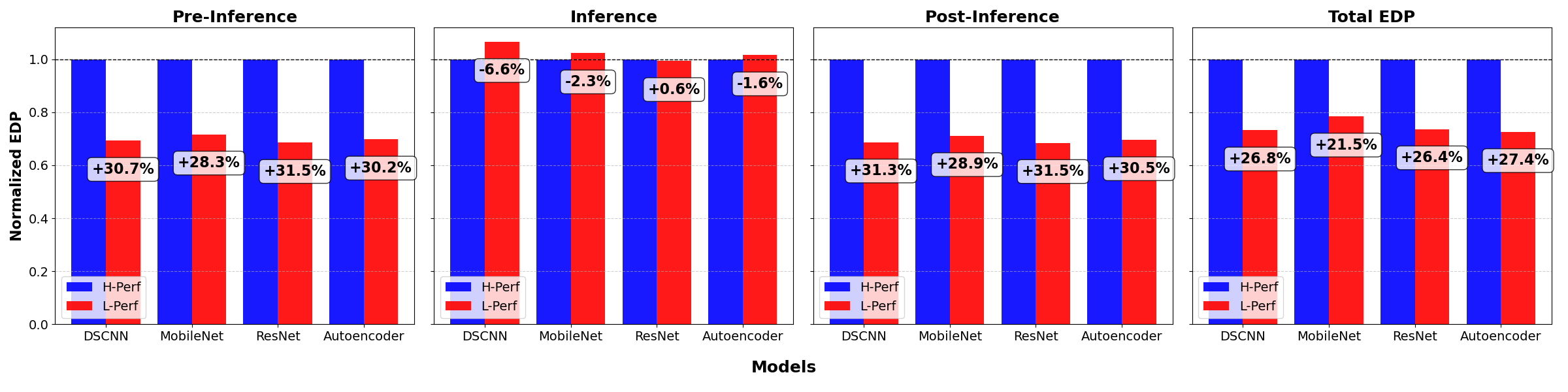}
    \captionsetup{font=small}
    \caption{Normalized EDP comparison between H-Perf and L-Perf across the three execution phases and total execution, with values normalized to H-Perf. L-Perf reduces EDP in pre- and post-inference, while inference shows a slight variation, with ResNet being the only model maintaining a positive rEDP.}
    \label{fig:bar}
\end{figure*}

Figure~\ref{fig:bar} illustrates how the rEDP is positive for both the pre- and post-inference phases, indicating improved energy efficiency in these stages. 
Conversely, for the inference phase, rEDP is slightly negative, except in the case of ResNet. 

Nevertheless, an overall decrease in EDP is observed when considering the total execution, demonstrating that the reduction in core voltage and clock frequencies effectively led to an improvement in energy efficiency across the entire process. 
This result highlights the importance of phase separation, as it allowed us to identify that reducing the core voltage was the best approach to increase energy efficiency.

\begin{table*}[]
    \centering
    \scriptsize
    \renewcommand{\arraystretch}{1.3}
    \setlength{\tabcolsep}{6pt}
    \caption{Comparison of \textbf{energy consumption} ($\mu J$) for different models under H-Perf and L-Perf conditions. Values are expressed as Mean ± Standard Deviation.}
    \resizebox{\textwidth}{!}{ 
    \begin{tabular}{l|cccc|cccc}
        \toprule
        & \multicolumn{4}{c|}{\textbf{H-Perf}} & \multicolumn{4}{c}{\textbf{L-Perf}} \\
        \cmidrule(lr){2-5} \cmidrule(lr){6-9}
        & Pre-Inference & Inference & Post-Inference & Total & Pre-Inference & Inference & Post-Inference & Total \\
        \midrule
        \textbf{DSCNN}  
        & $146.4 \pm 3.8$ & $34.1 \pm 0.7$ & $38.4 \pm 1.2$ & $219.0 \pm 4.1$ 
        & $100.8 \pm 3.7$ & $29.2 \pm 0.8$ & $26.5 \pm 1.1$ & $156.5 \pm 3.9$ \\
        \textbf{MobileNet}  
        & $267.8 \pm 5.8$ & $136.0 \pm 1.9$ & $40.2 \pm 1.3$ & $443.9 \pm 6.2$
        & $191.5 \pm 5.3$ & $111.4 \pm 2.1$ & $28.5 \pm 1.1$ & $331.4 \pm 5.8$ \\
        \textbf{ResNet}  
        & $162.3 \pm 4.2$ & $71.2 \pm 0.9$ & $43.5 \pm 1.5$ & $277.1 \pm 4.6$  
        & $111.1 \pm 3.4$ & $56.7 \pm 0.9$ & $29.7 \pm 1.1$ & $197.6 \pm 3.7$ \\
        \textbf{Autoencoder}  
        & $150.3 \pm 4.7$ & $24.9 \pm 0.6$ & $40.8 \pm 1.5$ & $216.1 \pm 4.9$  
        & $104.7 \pm 3.3$ & $20.4 \pm 0.6$ & $28.4 \pm 1.2$ & $153.6 \pm 3.6$ \\
        \bottomrule
    \end{tabular}
    }
    \label{tab:energy}
\end{table*}

\vspace{10pt} 

\begin{table*}[]
    \centering
    \scriptsize
    \renewcommand{\arraystretch}{1.3}
    \setlength{\tabcolsep}{6pt}
    \caption{Comparison of \textbf{model latency} ($\mu s$) for different models under H-Perf and L-Perf conditions. Values are expressed as Mean ± Standard Deviation.}
    \resizebox{\textwidth}{!}{ 
    \begin{tabular}{l|cccc|cccc}
        \toprule
        & \multicolumn{4}{c|}{\textbf{H-Perf}} & \multicolumn{4}{c}{\textbf{L-Perf}} \\
        \cmidrule(lr){2-5} \cmidrule(lr){6-9}
        & Pre-Inference & Inference & Post-Inference & Total & Pre-Inference & Inference & Post-Inference & Total \\
        \midrule
        \textbf{DSCNN}  
        & $1135.4 \pm 3.1$ & $155.3 \pm 0.2$ & $295.8 \pm 3.5$ & $1586.43 \pm 4.67$ 
        & $1136.0 \pm 7.4$ & $193.4 \pm 0.2$ & $295.9 \pm 3.7$ & $1625.3 \pm 8.2$ \\
        \textbf{MobileNet}  
        & $1992.0 \pm 8.5$ & $608.9 \pm 0.5$ & $324.4 \pm 3.8$ & $2925.3 \pm 9.3$
        & $1994.0 \pm 8.4$ & $760.6 \pm 0.7$ & $324.5 \pm 3.6$ & $3079.1 \pm 9.1$ \\
        \textbf{ResNet}  
        & $1228.7 \pm 13.5$ & $234.9 \pm 2.2$ & $331.0 \pm 6.8$ & $1794.5 \pm 15.3$  
        & $1229.9 \pm 12.9$ & $292.9 \pm 0.3$ & $331.4 \pm 6.5$ & $1854.1 \pm 14.5$ \\
        \textbf{Autoencoder}  
        & $1145.3 \pm 11.8$ & $128.9 \pm 2.8$ & $313.0 \pm 6.3$ & $1587.3 \pm 13.7$  
        & $1145.7 \pm 11.7$ & $160.2 \pm 2.9$ & $313.1 \pm 6.3$ & $1619.0 \pm 13.6$ \\
        \bottomrule
    \end{tabular}
    }
    \label{tab:time}
\end{table*}

\begin{table*}[]
    \centering
    \small 
    \renewcommand{\arraystretch}{1.1} 
    \setlength{\tabcolsep}{4pt} 
    \caption{Comparison of the Energy Delay Product (EDP) and relative EDP (rEDP) for different models. The models tested are: DSCNN, MobileNet, ResNet, and Autoencoder. H-Perf and L-Perf represent the two test setups used for the STM32N6.}
    \resizebox{\textwidth}{!}{ 
    \begin{tabular}{lcccc|cccc|cccc}
        \toprule
        & \multicolumn{8}{c|}{\textbf{EDP}} & \multicolumn{4}{c}{\textbf{rEDP}} \\
        \cmidrule(lr){2-9} \cmidrule(lr){10-13}
        & \multicolumn{4}{c|}{\textbf{H-Perf}} & \multicolumn{4}{c|}{\textbf{L-Perf}} & \multicolumn{4}{c}{\textbf{}} \\
        \cmidrule(lr){2-5} \cmidrule(lr){6-9} \cmidrule(lr){10-13}
        & Pre-Inf & Inf & Post-Inf & Total & Pre-Inf & Inf & Post-Inf & Total & Pre & Inf & Post & Tot \\
        \midrule
        DSCNN & $1.7 \times 10^{-7}$ & $5.3 \times 10^{-9}$ & $1.1 \times 10^{-8}$ & $3.5 \times 10^{-7}$ & 
                $1.1 \times 10^{-7}$ & $5.6 \times 10^{-9}$ & $7.8 \times 10^{-9}$ & $2.5 \times 10^{-7}$ & 
                31\% & -7\% & 31\% & 27\% \\
        MobileNet & $5.3 \times 10^{-7}$ & $8.3 \times 10^{-8}$ & $1.3 \times 10^{-8}$ & $1.3 \times 10^{-6}$ & 
                    $3.8 \times 10^{-7}$ & $8.5 \times 10^{-8}$ & $9.2 \times 10^{-9}$ & $1.0 \times 10^{-6}$ & 
                    28\% & -2\% & 29\% & 21\% \\
        Resnet & $2.0 \times 10^{-7}$ & $1.7 \times 10^{-8}$ & $1.4 \times 10^{-8}$ & $4.9 \times 10^{-7}$ & 
                 $1.4 \times 10^{-7}$ & $1.7 \times 10^{-8}$ & $9.9 \times 10^{-9}$ & $3.7 \times 10^{-7}$ & 
                 32\% & 1\% & 32\% & 26\% \\
        Autoencoder & $1.7 \times 10^{-7}$ & $3.2 \times 10^{-9}$ & $1.3 \times 10^{-8}$ & $3.4 \times 10^{-7}$ & 
                      $1.2 \times 10^{-7}$ & $3.3 \times 10^{-9}$ & $8.9 \times 10^{-9}$ & $2.5 \times 10^{-7}$ & 
                      30\% & -2\% & 30\% & 28\% \\
        \bottomrule
    \end{tabular}
    }
    \label{tab:redp}
\end{table*}

\section{Conclusion and Future Works}

This work presents a benchmarking methodology to improve the evaluation of machine learning performance on resource-constrained devices, such as MCUs. 
By addressing limitations in existing frameworks, our approach enables precise measurement of energy consumption, latency, and overall system efficiency.

A key aspect of our methodology is the EDP as a unified metric for assessing performance trade-offs in resource-constrained environments. 
By integrating energy and latency, EDP simplifies comparisons across configurations and platforms. 
While widely used in other contexts, we propose its adoption for benchmarking MCU architectures, particularly for NPU-equipped systems. 
To refine this approach, the rEDP proves especially useful for comparing energy efficiency across different architectures and configurations, providing a clearer assessment of performance trade-offs.

To enable precise profiling, our methodology incorporates a dual-trigger setup that isolates execution phases, allowing detailed analysis of latency and energy consumption. 
This facilitates optimal configuration selection and helps identify inefficient stages for targeted optimizations.
Our analysis pinpointed the pre-inference phase as the primary bottleneck, and optimizing it led to an average $25.5\%$ reduction in total EDP across all tested MLCommons models.

Additionally, our setup enables simultaneous measurement of energy, latency, and accuracy within the same firmware, ensuring consistent and reliable comparisons. 
The high level of automation improves statistical reliability while minimizing uncertainties.

The methodology was validated on the STM32N6 MCU. 
Our dual-trigger approach enables precise energy efficiency analysis, revealing opportunities for fine-tuning and optimization. 
This allows for detecting inefficiencies and implementing targeted improvements, improving energy efficiency.

Future work will focus on validating the proposed methodology across additional MCUs with diverse architectures and refining it by isolating and subtracting baseline energy contributions from non-essential components active during inference.
These enhancements will further optimize systems where inference is fully offloaded to an NPU.

\section*{Acknowledgements}
This work was carried out in the EssilorLuxottica Smart Eyewear Lab, a Joint Research Center between EssilorLuxottica and Politecnico di Milano. 
STMicroelectronics is acknowledged for providing early access to STM32N6 before its commercial release and relevant AI mapping tools, which facilitated the testing and benchmarking efforts underlying this work.

\bibliographystyle{IEEEtran}  
\bibliography{Bibliography}
\end{document}